\documentclass{article}

\usepackage{iclr2027_conference,times}

\usepackage[utf8]{inputenc}
\usepackage[T1]{fontenc}
\usepackage[table]{xcolor}
\usepackage{hyperref}
\usepackage{url}
\usepackage{booktabs}
\usepackage{amsfonts}
\usepackage{amsmath}
\usepackage{amssymb}
\usepackage{amsthm}
\usepackage{microtype}
\usepackage{graphicx}
\usepackage{wrapfig}

\usepackage{amsmath,amsfonts,bm}

\def\eqref#1{equation~\ref{#1}}

\def\1{\bm{1}}

\DeclareMathAlphabet{\mathsfit}{\encodingdefault}{\sfdefault}{m}{sl}
\SetMathAlphabet{\mathsfit}{bold}{\encodingdefault}{\sfdefault}{bx}{n}

\usepackage{enumitem}

\newtheorem{theorem}{Theorem}
\newtheorem{assumption}{Assumption}

\usepackage{enumitem}
\usepackage{tcolorbox}
\tcbuselibrary{breakable}

\newcommand{\method}{\textsc{Tips}}
\newcommand{\chg}[1]{\textcolor{gray}{\scriptsize(#1)}}
\newcommand{\makecell}[1]{\shortstack{#1}}

\title{Inducing Process Supervision from Outcome-Only Reinforcement Learning}

\author{
Shengda Fan$^{1}$\hspace{0.5em},
Xin Cong$^{2}$\hspace{0.5em},
Zhong Zhang$^{3}$\hspace{0.5em},
Haotian Chen$^{4}$\hspace{0.5em},
Yankai Lin\thanks{Corresponding author.}
\\
$^{1}$Renmin University of China \\
$^{2}$Tsinghua University \\
$^{3}$University of Electronic Science and Technology of China \\
$^{4}$Shanghai Jiao Tong University \\
\texttt{fanshengda@ruc.edu.cn, yankailin@ruc.edu.cn}
}

\iclrfinalcopy

\begin{document}

\maketitle

\fancyhead{} 
\renewcommand{\headrulewidth}{0pt}

\begin{abstract}
Process reward models (PRMs) have become a key component for LLMs, as their step-level feedback supports both post-training and test-time
reasoning. However, training strong PRMs remains costly: human step annotation is difficult to scale, while Monte Carlo estimation is computationally expensive and can drift from the intrinsic correctness of steps. To get effective PRMs at low cost, we introduce \method{} (\textbf{T}hinking-\textbf{I}nduced \textbf{P}rocess \textbf{S}upervision), an outcome-only reinforcement learning (RL) framework for training generative PRMs. In \method{}, the model generates a chain-of-thought (CoT) followed by step-level labels and an outcome label. The reward depends solely on whether the predicted outcome matches the ground truth, and the resulting group-relative advantage is used to optimize the entire generated response. Intuitively, when checking intermediate steps helps determine the outcome, more accurate checks can lead to better outcome judgments and higher rewards. Outcome-only RL can therefore reinforce step-level verification without explicit process supervision.
We validate the effectiveness of \method{} across math and
agent benchmarks and four backbone families. Notably,
\method{}-Qwen3-4B-Thinking-2507 reaches \textbf{85.2 F1} on ProcessBench with
only 3.2K outcome-labeled trajectories, surpassing all evaluated trained PRMs
and strong prompt-only judges such as GPT-5.4-Instruct and Claude-4.7-Opus,
while still trailing o1-mini. Code and data are
available at \url{https://github.com/RUCBM/TIPS}.
\end{abstract}

\begin{figure}[h]
    \centering
    \includegraphics[width=1\linewidth]{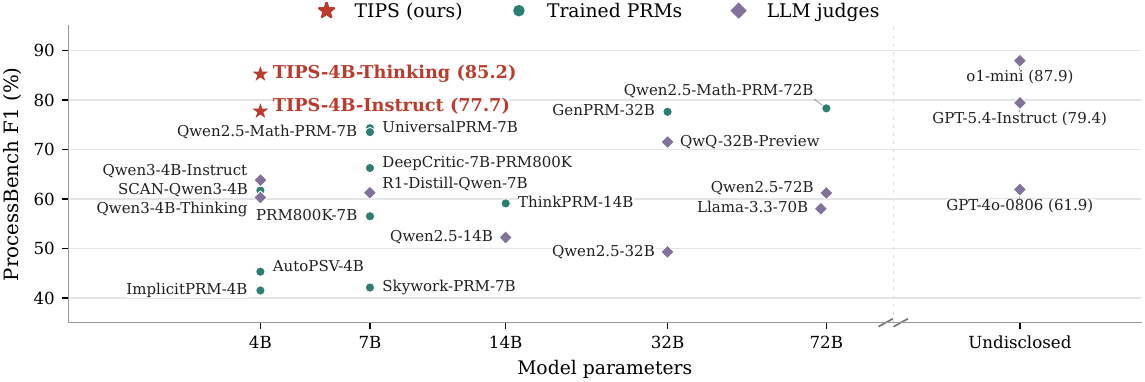}
    \caption{\textbf{ProcessBench performance across model scales.} \method{} achieves strong performance using a 4B-parameter model trained on only 3.2K publicly available, outcome-labeled reasoning traces.}
    \label{fig:placeholder}
\end{figure}

\section{Introduction}

Process reward models (PRMs), which provide step-level feedback for reasoning
traces and agent trajectories, have become a central component of test-time
scaling and fine-grained credit assignment in large language models
(LLMs)~\citep{lightman2023let,wang-etal-2024-math}. Despite their importance,
building strong PRMs remains difficult. Human step annotation provides reliable
supervision but is expensive and hard to scale~\citep{lightman2023let}. The
prevailing Monte Carlo bassed methods~\citep{ding2026scan, wang-etal-2024-math} reduce human cost, but require massive
rollouts and defines step quality through the existence of a correct
continuation.
However, the possibility of reaching a correct answer from a step does not necessarily imply that the step itself is correct~\citep{zhang-etal-2025-lessons}.
Therefore, developing strong PRMs at low cost remains an open challenge.
%

\begin{figure}
    \centering
    \includegraphics[width=1\linewidth]{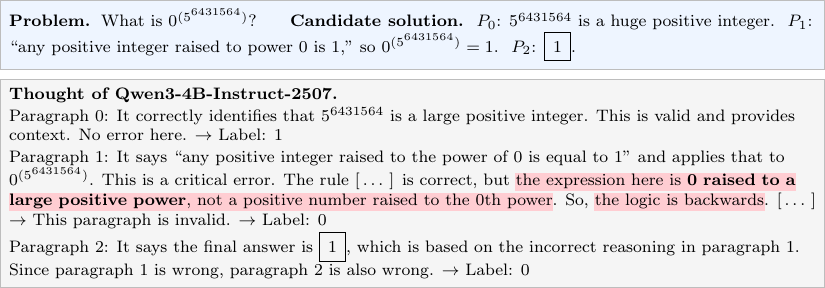}
    \caption{{Process verification emerges in generated thoughts.}}
    \label{fig:teaser}
\end{figure}

Our starting point is that outcome verification can benefit from process verification. When prompted as an outcome reward model (ORM), an LLM can inspect intermediate steps in its chain of thought (CoT) before judging the final answer. For example, Figure~\ref{fig:teaser} shows that Qwen3-4B identifies an erroneous step in its thought and consequently judges the final answer to be incorrect. This observation suggests that accurate step-level verification can support more reliable outcome judgments. It therefore raises a natural question: \textbf{can improving outcome verification, in turn, strengthen process verification?}

To investigate this question, we introduce \textbf{\method{}} (\textbf{T}hinking-\textbf{I}nduced \textbf{P}rocess \textbf{S}upervision), an outcome-only reinforcement learning framework for strengthening process verification through the model's generated thought. During training, 
as shown in Figure~\ref{fig:main_figure},
\method{} prompts an LLM to generate a CoT followed by step-level labels and an outcome label. The reward depends solely on the correctness of the outcome prediction. Under GRPO~\citep{shao2024deepseekmath}, the same outcome-derived advantage is applied to every token in the generated response, including those in the CoT, step labels, and outcome label.
Thus, outcome-level feedback shapes both the generated reasoning and the process judgments without direct process supervision.
Intuitively, rollouts that accurately evaluate step validity within their thinking process are more likely to predict the correct outcome, thereby earning higher relative advantages. 
Consequently, \textbf{rewarding correct outcome judgments can strengthen step-level verification in the model's thoughts}, even without process supervision.
Our theoretical analysis provides support for this intuition. When step validity determines the outcome and accurate outcome prediction requires the thought, that thought must contain information about step correctness.
We validate \method{} on both mathematical reasoning and agent tasks across
four backbone families: 
Qwen~\citep{yang2025qwen3, qwen2.5}, DeepSeek~\citep{deepseekai2025deepseekr1incentivizingreasoningcapability}, LlaMa~\citep{grattafiori2024llama, wang2025octothinker}, and SmolLM~\citep{bakouch2025smollm3}. The extensive experimental results demonstrate the effectiveness of \method{}. Notably, on widely used ProcessBench~\citep{zheng2025processbench}, \method{}-Qwen3-4B-Thinking-2507 achieves \textbf{85.2 F1} using only 3.2K outcome-labeled trajectories, outperforming substantially larger trained PRMs, as well as strong prompt-only proprietary judges.

In summary, our main contributions are:
\begin{itemize}[topsep=1pt, partopsep=1pt, itemsep=-1pt, leftmargin=10pt]
    \item We introduce \method{}, an outcome-only RL framework that computes rewards based solely on the correctness of outcome predictions and strengthens step-level verification as a byproduct.

    \item We provide an information-theoretic analysis of \method{}, deriving a lower bound on the information about step correctness encoded in the model's thought.

    \item We validate \method{} on mathematical reasoning and agent tasks across multiple backbones. 
\end{itemize}

\section{Related Work}

\paragraph{Process Reward Models.}
PRMs provide step-level supervision for LLM and have been shown to outperform outcome-only verification~\citep{lightman2023let}. To avoid costly human step
annotations, prior work commonly trains discriminative PRMs to fit automatically
constructed process labels, such as those derived from Monte Carlo estimation~\citep{ding2026scan, wang-etal-2024-math} or
stronger-model distillation~\citep{zhang-etal-2025-lessons,tan2026aurora}. However, these approaches typically use the LLM
only as a representation encoder with a scalar scoring head, leaving its
generative reasoning capability underexplored. Among these discriminative approaches, ImplicitPRM~\citep{yuan2025free} and
AutoPSV~\citep{lu2024autopsv} are closest to our supervision setting, as they
learn process scores from trajectory-level outcome labels; however, their
induced signals often fail to reliably localize step-level errors (Table~\ref{tab:main_result_processbench}).
Recent generative PRMs instead prompt models to produce chain-of-thought rationales before making step-level judgments~\citep{zhao2025genprm,khalifa2026process}, but still rely
on explicit process supervision from human annotated steps or stronger judges. In
contrast, \method{} uses only trajectory-level outcome labels and induces
step-level verification through outcome-only RL.

\paragraph{Reinforcement Learning for LLMs.}
Since the success of
DeepSeek-R1~\citep{deepseekai2025deepseekr1incentivizingreasoningcapability},
reinforcement learning has become an increasingly important post-training
paradigm for LLMs. Recent studies have applied RL to improve LLMs in
mathematical reasoning~\citep{yu2026dapo, fan2026darc}, tool-using
agents~\citep{li2025torl, fan2026generalizing}, retrieval~\citep{jin2025search,chen2026agentcpm},
code generation~\citep{wang2025rlcoder, team2025minicpm4}, demonstrating its effectiveness. Beyond policy
model learning, a growing line of work explores RL for reward models~\citep{chen2026rmr,whitehouse2025j1,guo2025reward},
showing that RL can improve ORM accuracy. Our work differs in its objective:
rather than using RL only to obtain a stronger ORM, we show that
outcome-only RL can induce a strong PRM as a by-product.

\section{Thinking-Induced Process Supervision}

In this section, we present \method{}, a framework that induces process supervision from outcome-only reinforcement learning. We first motivate the key intuition (Section~\ref{sec:motivation}), then describe the training framework (Section~\ref{sec:method}), and finally provide an information-theoretic analysis (Section~\ref{sec:analysis}).

\subsection{Motivation}\label{sec:motivation}

Recent work has shown that prompting LLMs to think before acting can improve performance~\citep{wei2022chain,yao2022react}. We observe a related
phenomenon in reward modeling: when prompted as an ORM,
an LLM often uses its thought to inspect intermediate steps before producing the
final verdict. As illustrated in Fig.~\ref{fig:teaser}, such thoughts already
contain useful step-level verification even without process-level training.
This suggests a simple way to induce process supervision from outcome-only
feedback. If a thought identifies step-level errors more faithfully, the model is
more likely to predict the correct trajectory-level outcome; under RL, such
thought patterns receive higher reward and are reinforced. \textbf{Thus, outcome-only
optimization can indirectly strengthen process-aware verification, without
requiring explicit step annotations.}
Based on this insight, we propose \method{}, a simple framework that trains a
generative reward model with outcome-only RL and obtains strong PRM capability as
a byproduct.

\subsection{Methodology}\label{sec:method}

\begin{figure}
    \centering
    \includegraphics[width=1\linewidth]{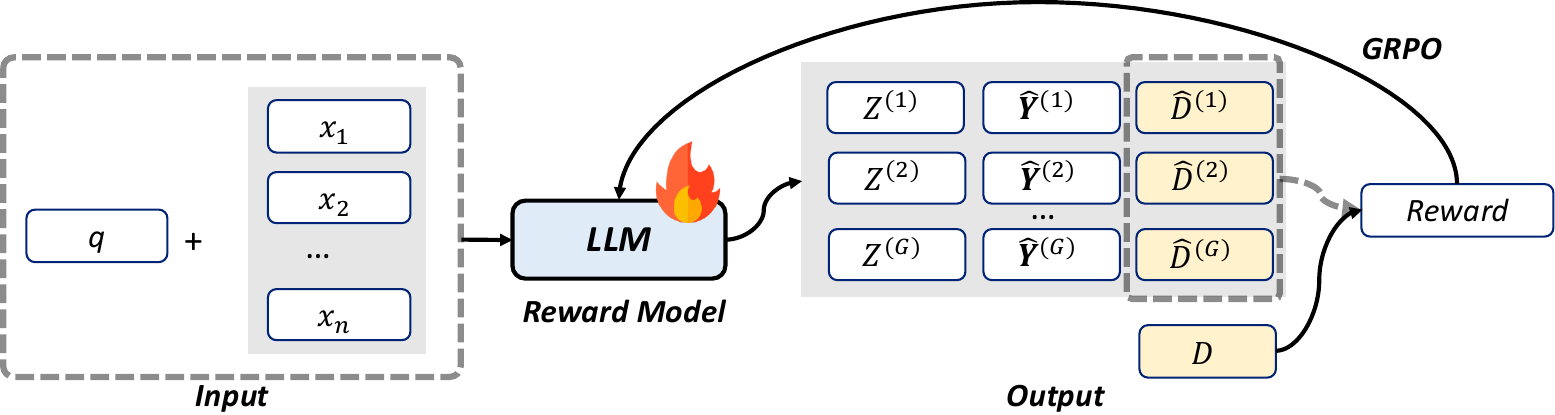}
    \caption{\textbf{Overview of \method{}.}
    Given an input trajectory, the model first generates a thinking chain and
    then predicts both process labels and an outcome label.
    During RL, only the outcome-level prediction is used to compute the reward,
    while the generated process labels receive no direct supervision.}
    \label{fig:main_figure}
\end{figure}

As shown in Fig.~\ref{fig:main_figure}, \method{} trains a generative reward
model to produce both outcome-level and step-level judgments, while using only
outcome-level ground truth for supervision. Given a problem $q$ and a candidate
response $\mathbf{x}=(x_1,\dots,x_n)$ with $n$ intermediate steps, we prompt the
model to first generate a thinking chain
$Z \sim \pi_\theta(\cdot \mid q,\mathbf{x})$, where $\pi_\theta$ denotes the generative reward model. Conditioned on this thought, the
model then outputs two types of judgments: a sequence of process labels
$\boldsymbol{\hat{Y}}=(\hat{Y}_1,\dots,\hat{Y}_n)$, where
$\hat{Y}_t\in\{0,1\}$ indicates whether step $x_t$ is valid given the preceding
steps, and an outcome label $\hat{D}\in\{0,1\}$ predicting whether the full
trajectory is correct.
Let $D\in\{0,1\}$ denote the ground-truth outcome label. We define the reward
using only the outcome prediction:
\begin{equation}
r =
\begin{cases}
1, & \text{if } \hat{D}=D, \\
0, & \text{otherwise}.
\end{cases}
\label{eq:outcome_reward}
\end{equation}
Importantly, no process label is used in the reward.
We optimize the reward model with GRPO~\citep{shao2024deepseekmath}, a
critic-free PPO-style algorithm~\citep{schulman2017proximal}.
For each input $(q, \mathbf{x})$, we sample a group of $G$ rollouts $\{(Z^{(i)}, \smash{\boldsymbol{\hat{Y}}^{(i)}}, \hat{D}^{(i)})\}_{i=1}^G$ and compute a group-relative advantage based on the rewards $\{r^{(i)}\}_{i=1}^G$ defined in Eq.~\ref{eq:outcome_reward}.
Rollouts that produce the correct outcome verdict receive higher relative advantage and are
reinforced. Therefore, when correct outcome prediction depends on tracking step
validity, outcome-only RL selects for thoughts that perform more faithful
process verification. We provide the full GRPO objective and implementation
details in Appendix~\ref{app:grpo}.
At inference time, we use the generated process labels $\boldsymbol{\hat{Y}}$ directly as
step-level labels. Thus, \method{} requires only outcome-labeled
trajectories for training, but yields a PRM-style verifier at test time.

\subsection{Analysis}\label{sec:analysis}

We provide an information-theoretic account of why outcome-only optimization induces process supervision. We consider the random variables $(Q, \mathbf{X}) \sim \mathcal{D}$ corresponding to the instances $(q, \mathbf{x})$ defined in Sec.~\ref{sec:method}. Let $\mathbf{Y}=(Y_1, \dots, Y_n)$ and $D \in \{0,1\}$ denote the ground-truth step-level labels and outcome label, respectively. Let
$R=\mathrm{enc}_\theta(Q,\mathbf{X})$ denote the model's pre-reasoning
representation, and let $Z\sim\pi_\theta(\cdot\mid Q,\mathbf{X})$ denote the
generated reasoning trace used to produce the final verdict. We write
$\hat D=g(R,Z)$ for the parsed outcome prediction. For the analysis, we treat
the parser $g$ as deterministic, so the randomness of $\hat D$ comes only from
the sampled trace $Z$. All mutual information quantities are taken over the
joint distribution induced by $\mathcal{D}$ and $\pi_\theta$.

\begin{assumption}[Nontrivial Verification]
\label{asm:nontrivial}
The pre-reasoning representation does not fully determine the outcome label:
there exists $\varepsilon>0$ such that
\begin{equation}
    H(D\mid R) \ge \varepsilon .
\end{equation}
\end{assumption}

This rules out the degenerate case where the model can infer the final verdict
directly from $R$, which is natural in nontrivial verification settings
where a single-pass representation is insufficient~\citep{li2024chain}.


\begin{theorem}[Outcome Information Lower Bound]
\label{thm:outcome}
Let $P_e=\mathbb{P}[\hat D\neq D]$. Then
\begin{equation}
    I(Z;D\mid R)
    \ge
    H(D\mid R)-H_b(P_e),
\end{equation}
where $H_b(\cdot)$ is the binary entropy function.
\end{theorem}

By Fano's inequality~\citep{cover2012elements}, low outcome error implies that
the thinking chain must explain a nontrivial fraction of the uncertainty about
$D$ that remains after observing $R$.
We next relate this outcome information to process information. The key quantity
is the residual uncertainty $H(D\mid \mathbf{Y},R)$, which measures how much the
outcome remains undetermined even after the step-level labels are known.

\begin{theorem}[Process Information Lower Bound]
\label{thm:process}
For arbitrary process labels $\mathbf{Y}$,
\begin{equation}
    I(Z;\mathbf{Y}\mid R)
    \ge
    I(Z;D\mid R)-H(D\mid \mathbf{Y},R).
\end{equation}
\end{theorem}

Theorem~\ref{thm:process} shows that outcome-relevant information in $Z$ must
also reveal process information when the outcome is largely determined by the
step-level correctness pattern. This condition naturally holds in mathematical
reasoning: once the validity of the intermediate steps is known, the correctness
of the final answer is often nearly determined, making
$H(D\mid \mathbf{Y},R)$ small.

The proof of the above theorems can be found in Appendix~\ref{app:thinking_process_supervision}. Combining Theorems~\ref{thm:outcome} and~\ref{thm:process} gives
\begin{equation}
\label{eq:combined}
    I(Z;\mathbf{Y}\mid R)
    \ge
    \underbrace{H(D\mid R)}_{\text{verification difficulty}}
    -
    \underbrace{H_b(P_e)}_{\text{outcome error}}
    -
    \underbrace{H(D\mid \mathbf{Y},R)}_{\text{step--outcome residual}} .
\end{equation}

Eq.~\ref{eq:combined} captures the central mechanism behind \method{}.
When verification is nontrivial, outcome error is low, and the final outcome is
tightly coupled with step-level correctness, the generated thought must
encode information about the reasoning process. This also explains why
outcome-only GRPO can favor process-aware thoughts. For a fixed input, GRPO
samples multiple rollouts with different CoT and outcome
predictions. Rollouts that produce the correct outcome judgment receive higher
relative advantage and are reinforced. When correct outcome prediction depends
on tracking step validity, this reward signal implicitly selects for thoughts
that encode process-relevant information, even though the reward never directly
observes step-level labels.

The same analysis identifies when the transfer should weaken. \textbf{First}, input-level shortcuts
weaken the first term: if the outcome can already be predicted from $R$, then
$H(D\mid R)$ is small and the thought channel need not encode process
information. \textbf{Second}, ineffective outcome optimization weakens the second term:
if RL fails to reduce the outcome error $P_e$, then $H_b(P_e)$ remains large and
the lower bound in Eq.~\ref{eq:combined} becomes loose. \textbf{Third}, loose
step--outcome coupling weakens the third term: when step labels only indirectly
determine task success, $H(D\mid\mathbf{Y},R)$ is large, so outcome information
need not translate into process information. \textbf{Fourthly}, even when useful
information is required beyond $R$, a weak backbone may fail to express it as
faithful step-by-step reasoning. We
examine these regimes empirically in Sec.~\ref{sec:discussions}.

\section{Experiments}

\subsection{Experimental Setup}\label{sec:exp_setup}

\paragraph{Models.} To evaluate the generalization of \method\  across different backbone models, we conduct experiments using both instruct and thinking models of various scales. The models we experiment include Qwen~\citep{yang2025qwen3, qwen2.5}, DeepSeek~\citep{deepseekai2025deepseekr1incentivizingreasoningcapability}, LlaMa~\citep{grattafiori2024llama, wang2025octothinker}, and SmolLM~\citep{bakouch2025smollm3} families.
The implementation details can be found in Appendix~\ref{app:imp_details}.

\paragraph{Benchmarks and Metrics.}
We evaluate \method\ in two domains: mathematical reasoning and LLM-agent tasks.
For mathematical reasoning, we first adopt the standard \textbf{Best-of-$8$}~\citep{lightman2023let} evaluation to assess whether a PRM can improve downstream response selection.
Given eight sampled responses from Qwen2.5-7B-Instruct for each problem, each PRM selects the response with the highest score, and we report the selected-answer accuracy on GSM8K~\citep{cobbe2021gsm8k}, MATH~\citep{hendrycks2021measuring}, College Math~\citep{tangmathscale}, Minerva~\citep{lewkowycz2022solving}, and OlympiadBench~\citep{he2024olympiadbench}.
To ensure a fair comparison, all PRMs are evaluated on the same set of sampled responses.
We additionally report Majority Vote@8 as a reference baseline and Pass@8 as an oracle upper bound.
Beyond response selection, we further evaluate the PRMs' ability to detect step-level errors on ProcessBench~\citep{zheng-etal-2025-processbench}, where models are required to identify the first erroneous step or determine that all steps are correct. For ProcessBench, we report the official F1 score. Finally, we evaluate on AgentProcessBench~\citep{fan2026agentprocessbench}, where PRMs assign labels to assistant responses in multi-turn agent trajectories. We report the benchmark's official First-Error Accuracy (FirstErrAcc).

\paragraph{Baselines.}
We compare \method\ with four groups of baselines, organized by the form and cost of their supervision.
\textbf{(1) LLM-as-a-Judge} baselines use the open-source backbones as well as stronger proprietary GPT and Claude models as prompt-only critics, testing whether reliable process evaluation can be obtained without PRM training.
\textbf{(2) Published discriminative PRMs} include representative step-wise reward models such as Math-Shepherd~\citep{wang-etal-2024-math}, Qwen2.5-Math-PRM~\citep{zhang2025lessons}, RLHFlow-PRM~\citep{dong2024rlhf}, EurusPRM~\citep{cui2025process}, UniversalPRM~\citep{tan2026aurora}, and Scan-PRM~\citep{ding2026scan}, whose supervision comes from Monte-Carlo-estimated step labels, stronger-model-distilled annotations, human process labels, or their combinations.
\textbf{(3) Published generative PRMs} include GenPRM~\citep{zhao2025genprm} and ThinkPRM~\citep{khalifa2026process}, which cast process evaluation as conditional generation rather than scalar scoring. Although they share the same formulation as \method, their supervision depends on 32B-level thinking LLMs to generate rationales, followed by costly filtering with Monte Carlo estimates or human step annotations.
\textbf{(4) Outcome-supervised implicit PRMs} include AutoPSV~\citep{lu2024autopsv} and ImplicitPRM~\citep{yuan2025free}, which learn process evaluators from trajectory-level outcome labels only. These methods form the closest weak-supervision baselines to \method, while \method\ differs by explicitly verbalizing process rationale.
Note that the published off-the-shelf PRMs are trained on different backbones with different data.
For a fair comparison among outcome-supervised implicit PRMs, we re-implement AutoPSV and ImplicitPRM under the same setting as \method. We also re-implement SCAN-PRM, a representative state-of-the-art Monte-Carlo-based PRM, as a strong supervised reference baseline.

\subsection{Results on Math}

\begin{table*}[t]
  \centering
  \scriptsize
  \setlength{\tabcolsep}{4.5pt}
  \caption{
  \textbf{Evaluation results of F1 on ProcessBench} (\%).
  \texttt{Human}: human-annotated step labels;
  \texttt{MC}: pseudo process labels obtained via Monte Carlo estimation;
  \texttt{Distill}: stronger-model-derived supervision;
  \texttt{Outcome}: trajectory-level outcome labels only.
  }
  \label{tab:main_result_processbench}
  \resizebox{\textwidth}{!}{%
  \begin{tabular}{@{}llccccc@{}}
  \toprule
  Model & Supervision & GSM8K & MATH & \makecell{Olympiad \\ Bench} & OmniMath & Avg. \\
  \midrule

  \multicolumn{7}{c}{\textbf{LLM-as-a-Judge}} \\
  \midrule
   Qwen3-4B-Instruct-2507 & N/A & 64.8 & 68.7 & 63.6 & 58.1 & 63.8 \\
    Qwen3-4B-Thinking-2507 & N/A & 52.1 & 70.1 & 59.5 & 59.4 & 60.3 \\

  \rowcolor{gray!15}
  GPT-5.4-Instruct & N/A & 89.8 & 84.6 & 74.1 & 69.1 & 79.4 \\
  \rowcolor{gray!15}
  Claude-4.7-Opus & N/A & 85.3 & 82.9 & 75.7 & 73.4 & 79.3 \\
  \rowcolor{gray!15}
  o1-mini & N/A & 93.2 & 88.9 & 87.2 & 82.4 & 87.9 \\

  \midrule
  \multicolumn{7}{c}{\textbf{Published Discriminative PRMs}} \\
  \midrule
  UniversalPRM-7B & Distill & 85.8 & 77.7 & 67.6 & 66.4 & 74.3 \\
   Qwen2.5-Math-PRM-72B & MC+Distill & \underline{87.3} & 80.6 & 74.3 & 71.1 & \underline{78.3} \\

  RLHFlow-PRM-Deepseek-8B & MC & 38.8 & 33.8 & 16.9 & 16.9 & 26.6 \\
  Qwen2.5-Math-7B-Math-Shep & MC & 62.5 & 31.6 & 13.7 & 7.7 & 28.9 \\
EurusPRM-Stage1 & Outcome & 54.7 & 41.2 & 24.7 & 17.5 & 34.5 \\
      EurusPRM-Stage2 & Outcome+Distill & 67.0 & 53.2 & 35.4 & 30.7 & 46.6 \\

  Qwen2.5-Math-7B-PRM800K & Human & 68.2 & 62.6 & 50.7 & 44.3 & 56.5 \\

  \midrule
  \multicolumn{7}{c}{\textbf{Published Generative PRMs}} \\
  \midrule
ThinkPRM-14B & Human+Distill & 65.3 & 64.0 & 53.9 & 52.9 & 59.1 \\

  GenPRM-32B & MC+Distill & 83.1 & {81.7} & 72.8 & \underline{72.8} & 77.6 \\
  \midrule
  \multicolumn{7}{c}{\textbf{Matched-Data Qwen3-4B Baselines}} \\
  \midrule
  SCAN-Qwen3-4B-Instruct-2507 & MC & 81.3 & 68.7 & 52.0 & 45.0 & 61.7 \\
  AutoPSV-Qwen3-4B-Instruct-2507 (3.2K) & Outcome & 39.0 & 33.4 & 20.3 & 21.5 & 28.6 \\
  AutoPSV-Qwen3-4B-Instruct-2507 (197K) & Outcome & 63.6 & 48.4 & 34.3 & 34.8 & 45.3 \\
  ImplicitPRM-Qwen3-4B-Instruct-2507 (3.2K) & Outcome & 52.6 & 30.8 & 9.1 & 14.4 & 26.7 \\
  ImplicitPRM-Qwen3-4B-Instruct-2507 (197K) & Outcome & 56.8 & 44.8 & 30.3 & 34.0 & 41.5 \\

  \midrule
  \multicolumn{7}{c}{\textbf{Ours}} \\
  \midrule
  \method-Qwen3-4B-Instruct-2507 (3.2K) & Outcome & 84.6 & \underline{82.6} & \underline{74.5} & 69.0 & 77.7 \\
  \method-Qwen3-4B-Thinking-2507 (3.2K) & Outcome & \textbf{89.5} & \textbf{87.7} & \textbf{83.5} & \textbf{80.0} & \textbf{85.2} \\
  \bottomrule
  \end{tabular}%
  }
\end{table*}

Table~\ref{tab:main_result_processbench} presents the results on ProcessBench, evaluating whether models can identify
step-level errors in reasoning trajectories. We highlight four observations.

\begin{itemize}[topsep=1pt, partopsep=1pt, itemsep=-1pt, leftmargin=10pt]

\item \textbf{Prior outcome-supervised PRMs do not induce reliable process labels.}
As shown in Table~\ref{tab:main_result_processbench}, prior outcome-supervised PRMs are weak step-level verifiers. Specifically, AutoPSV-$197$K and ImplicitPRM-$197$K achieve only $45.3$ and $41.5$ F1 on
ProcessBench, respectively.
This suggests that their preferences are not
grounded in explicit analysis of intermediate reasoning steps.

\item \textbf{\method\ turns outcome-only supervision into process-aware verification.}
Despite using only $3.2$K outcome-labeled trajectories,
\method-Qwen3-4B-Instruct-2507 achieves \textbf{$77.7$} average F1 on
ProcessBench, substantially outperforming AutoPSV-$197$K and
ImplicitPRM-$197$K. This suggests that the improvement comes not from simply
scaling outcome data, but from the explicit CoT used in \method, which
allows outcome supervision to be converted into process-level evidence. This is
consistent with Eq.~\ref{eq:combined}, since in mathematical reasoning the
final outcome is often largely determined by intermediate-step correctness.

\item \textbf{Thinking models amplify the emergence of process verification.}
When directly prompted as a judge, Qwen3-4B-Thinking-2507 is slightly weaker
than Qwen3-4B-Instruct-2507 on ProcessBench. After training with \method,
however, the thinking backbone reverses this gap and improves the average F1
from $77.7$ to $85.2$. This observation is consistent with our analysis in
Sec.~\ref{sec:analysis}: thinking models generate longer thoughts that encode
richer step-level information, which can be further reinforced during RL and
lead to stronger process verification.

\item \textbf{\method\ achieves the strongest performance among evaluated trained PRMs.}
With only a 4B-scale backbone, \method-Qwen3-4B-Thinking-2507 achieves
\textbf{85.2} average F1 on ProcessBench, outperforming previous state-of-the-art baselines, including Qwen2.5-Math-PRM-72B and GenPRM-32B, and proprietary models like GPT-5.4-Instruct and Claude-4.7-Opus, while still trailing o1-mini.
\end{itemize}

\subsection{Results on Agent}

Table~\ref{table:firsterracc_main} reports FirstErrAcc on AgentProcessBench.
We highlight two observations.

\begin{itemize}[topsep=1pt, partopsep=1pt, itemsep=-1pt, leftmargin=10pt]

\item \textbf{Outcome-only RL induces step-level localization beyond mathematics.}
On AgentProcessBench, \method\ improves average FirstErrAcc for all three
backbones, showing that the effect is not limited to mathematical reasoning. Same to the math domain, the gains are larger for thinking models: Qwen3-4B-Thinking-2507 and
Qwen3-8B-Thinking improve by $+5.4$ and $+4.5$ average FirstErrAcc, respectively, compared
with $+2.5$ for Qwen3-4B-Instruct-2507. This supports our central hypothesis:
when a model is rewarded only for the final judgment, RL can still reinforce
intermediate verification behaviors that help produce that judgment.

\item \textbf{Agent tasks exhibit weaker outcome--process coupling.}
AgentProcessBench yields smaller and more heterogeneous gains than ProcessBench. This is because of  weaker outcome--process coupling. In math, judging the final answer typically
requires checking each derivation step. In agent tasks, failures may instead
come from missing actions, insufficient evidence, or unproductive exploration,
which are not always tied to a clearly invalid observed step. Outcome feedback is
therefore less localized, making process supervision harder to induce.

\end{itemize}

\begin{table}[t]
    \centering
    \scriptsize
    \setlength{\tabcolsep}{3pt}
    \renewcommand{\arraystretch}{1.08}

    \caption{Comparison of FirstErrAcc on AgentProcessBench (\%).}
    \label{table:firsterracc_main}

    \begin{tabular}{
        l
        *{5}{r@{\;}l}
    }
    \toprule
    \textbf{Model}
    & \multicolumn{2}{c}{\textbf{HotPotQA}}
    & \multicolumn{2}{c}{\textbf{GAIA}}
    & \multicolumn{2}{c}{\textbf{BFCL}}
    & \multicolumn{2}{c}{\textbf{$\tau^2$-Bench}}
    & \multicolumn{2}{c}{\textbf{Avg.}} \\
    \midrule

    Qwen3-4B-Instruct-2507
    & 57.2 & {}
    & 35.6 & {}
    & 34.8 & {}
    & 42.4 & {}
    & 42.5 & {} \\
    \quad w/ \method
    & 58.0 & \chg{+0.8}
    & 38.0 & \chg{+2.4}
    & 35.6 & \chg{+0.8}
    & 48.4 & \chg{+6.0}
    & 45.0 & \chg{+2.5} \\

    \midrule

    Qwen3-4B-Thinking-2507
    & 60.4 & {}
    & 33.6 & {}
    & 37.2 & {}
    & 43.6 & {}
    & 43.7 & {} \\
    \quad w/ \method
    & 61.2 & \chg{+0.8}
    & 41.2 & \chg{+7.6}
    & 43.6 & \chg{+6.4}
    & 50.4 & \chg{+6.8}
    & 49.1 & \chg{+5.4} \\

    \midrule

    Qwen3-8B-Thinking
    & 56.8 & {}
    & 31.6 & {}
    & 30.8 & {}
    & 46.8 & {}
    & 41.5 & {} \\
    \quad w/ \method
    & 53.6 & \chg{-3.2}
    & 48.8 & \chg{+17.2}
    & 32.4 & \chg{+1.6}
    & 49.2 & \chg{+2.4}
    & 46.0 & \chg{+4.5} \\

    \bottomrule
    \end{tabular}
\end{table}

\begin{table*}[!t]
      \centering
      \scriptsize
      \setlength{\tabcolsep}{4.5pt}
      \caption{
      Best-of-8 reranking evaluation with Qwen2.5-7B-Instruct as the policy model.
      }
      \label{tab:best-of-8}
      \resizebox{\textwidth}{!}{%
      \begin{tabular}{@{}llccccc c@{}}
      \toprule
      Model & Supervision & GSM8K & MATH & \makecell{College \\ Math} & \makecell{Olympiad \\ Bench} & Minerva & Avg. \\
      \midrule

      \multicolumn{8}{c}{\textbf{Baselines and Upper Bound}} \\
      \midrule
      Majority Vote@8 & N/A & 93.8 & 80.1 & 66.4 & 45.9 & 52.6 & 67.8 \\
      \textbf{Pass@8 (Upper Bound)} & Oracle & 97.2 & 89.0 & 74.5 & 62.7 & 66.9 & 78.1 \\

      \midrule
      \multicolumn{8}{c}{\textbf{LLM-as-a-Judge}} \\
      \midrule
      Qwen3-4B-Instruct-2507 & N/A & 95.1 & 85.2 & 69.1 & 52.6 & 56.3 & 71.6 \\
      Qwen3-4B-Thinking-2507 & N/A & \underline{95.0} & \textbf{87.5} & \textbf{69.5} & \underline{56.2} & 55.9 & \underline{72.8} \\

      \midrule
      \multicolumn{8}{c}{\textbf{Matched-Data Qwen3-4B Baselines}} \\
      \midrule
      SCAN-PRM-Qwen3-4B-Instruct-2507 (197K) & MC & 94.8 & 82.1 & 68.1 & 48.3 & 52.6 & 69.2 \\
      AutoPSV-Qwen3-4B-Instruct-2507 (3.2K) & Outcome & 93.3 & 77.6 & 65.7 & 43.4 & 49.6 & 65.9 \\
      AutoPSV-Qwen3-4B-Instruct-2507 (197K) & Outcome & 94.2 & 79.0 & 66.5 & 43.9 & 53.3 & 67.4 \\
      ImplicitPRM-Qwen3-4B-Instruct-2507 (3.2K) & Outcome & 93.8 & 78.4 & 66.8 & 43.3 & 50.4 & 66.5 \\
      ImplicitPRM-Qwen3-4B-Instruct-2507 (197K) & Outcome & 94.2 & 81.8 & 68.4 & 46.2 & 53.3 & 68.8 \\

      \midrule
      \multicolumn{8}{c}{\textbf{Ours}} \\
      \midrule
      \method-Qwen3-4B-Instruct-2507 (3.2K) & Outcome & \underline{95.0} & \underline{86.9} & \underline{69.4} & 55.4 & \textbf{57.4} & \underline{72.8} \\
      \method-Qwen3-4B-Thinking-2507 (3.2K) & Outcome & \textbf{95.5} & \underline{87.4} & \textbf{69.5} & \textbf{56.7} & \textbf{57.4} & \textbf{73.3} \\
      \bottomrule
      \end{tabular}%
      }
  \end{table*}

\subsection{Discussions}\label{sec:discussions}

\paragraph{Best-of-$8$ Evaluation.}
Table~\ref{tab:best-of-8} reports Best-of-$8$ reranking accuracy with
Qwen2.5-7B-Instruct as the policy model. 
For \method, we first filter the candidate trajectories predicted to have correct final outcomes and then selects the retained trajectory with the highest mean predicted step-validity score. For baselines, we follow their default settings.
We highlight two observations.
\begin{itemize}[topsep=1pt, partopsep=1pt, itemsep=-1pt, leftmargin=10pt]
    \item \textbf{Strong prompt-only judges can outperform prior state-of-the-art trained PRMs.}
Without any PRM training, Qwen3-4B-Thinking-2507 achieves $72.8$ average accuracy, outperforming
the state-of-the-art MC-supervised \textsc{Scan}-PRM with $69.2$ and the
outcome-supervised ImplicitPRM with $68.8$.
This suggests that generative judges are already strong process-aware rerankers even without PRM training.
    \item \textbf{\method\ further improves generative PRMs with much lower supervision cost.}
    Using only $3.2$K outcome-only trajectories, \method-Qwen3-4B-Thinking-2507 reaches
    $73.3$ Avg, surpassing the $197$K-scale ImplicitPRM baseline by $+4.5$ points.
    This demonstrates the data efficiency of our framework in converting outcome-only
    supervision into stronger process evaluation ability. Meanwhile, \method\ still trails
    the Pass@$8$ oracle by $4.8$ points, with $73.3$ Avg compared to $78.1$ for Pass@$8$,
    leaving further headroom within the same candidate set.
\end{itemize}

\begin{figure}
    \centering
    \includegraphics[width=0.9\linewidth]{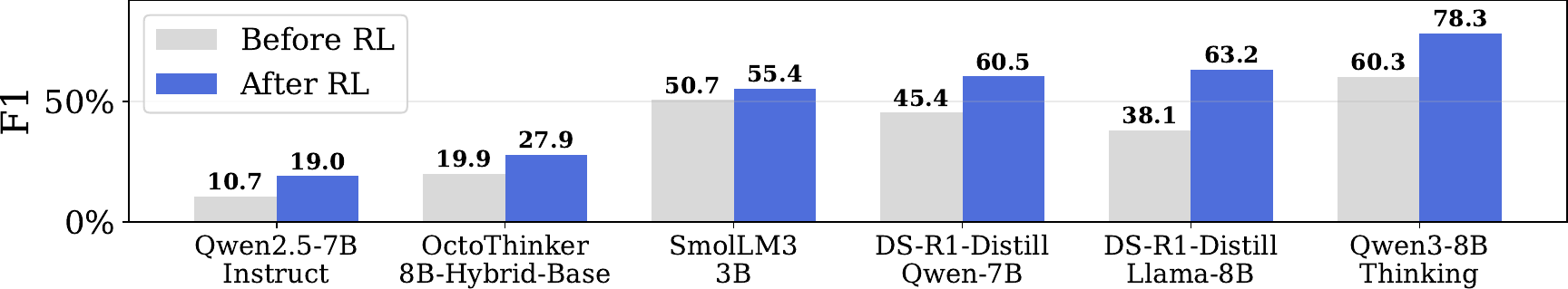}
    \caption{{\method\ improves ProcessBench F1 across diverse backbones.}}
\label{fig:generality}
\end{figure}

\paragraph{\method\ generalizes across model families and scales.}
To test whether \method\ is specific to the Qwen-2507-4B series, we run
additional math-domain experiments on diverse backbones. As shown in
Figure~\ref{fig:generality}, \method\ improves ProcessBench F1 across
DeepSeek-R1-distilled models, Llama-based models, SmolLM, Qwen2.5-7B-Instruct and Qwen3-8B. This
indicates that the gains are not tied to a single backbone family, but extend to
models with different architectures and scales.
In Appendix~\ref{app:solver_verifier}, we examine whether these gains can be explained solely by improved problem-solving ability.

\begin{figure}
    \centering
    \includegraphics[width=1\linewidth]{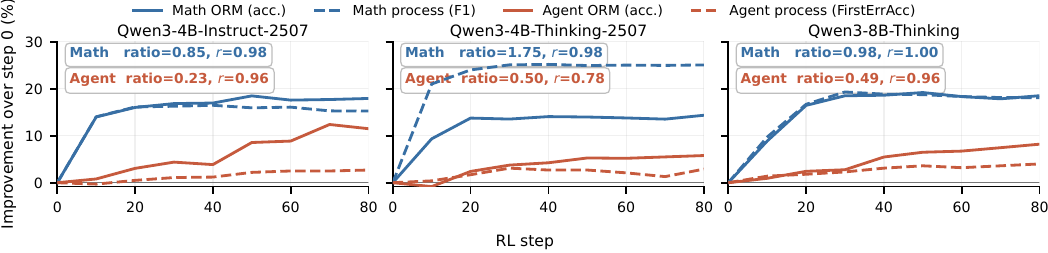}
\caption{\textbf{Outcome--process coupling during outcome-supervised RL.}
Solid lines show the absolute improvement in ORM accuracy, while dashed lines show the absolute improvement
in the corresponding PRM metric. All values are measured
relative to the base model. The annotated gain ratio is defined as
$\Delta_{\mathrm{process}} / \Delta_{\mathrm{ORM}}$, where each $\Delta$ is the
endpoint improvement. Pearson $r$ denotes the
correlation between ORM accuracy and the corresponding process metrics.}
\label{fig:orm_prm}
\end{figure}

\paragraph{Outcome--process coupling during outcome-supervised RL.}
We analyze the training dynamics of three backbones trained only with outcome rewards. As shown in Figure~\ref{fig:orm_prm}, ORM accuracy and process-level
metrics generally improve together, with Pearson correlations ranging from
$0.78$ to $1.00$. This suggests that outcome optimization can implicitly induce
process-level verification.
Meanwhile, we observe that math tasks show larger gain ratios, indicating more direct transfer from outcome judgment to step-level verification, while agent tasks show weaker transfer. This is likely because, in agent settings, the relation between local step correctness and final task success is more indirect.

\begin{figure}[t]
    \centering
    \includegraphics[width=0.9\linewidth]{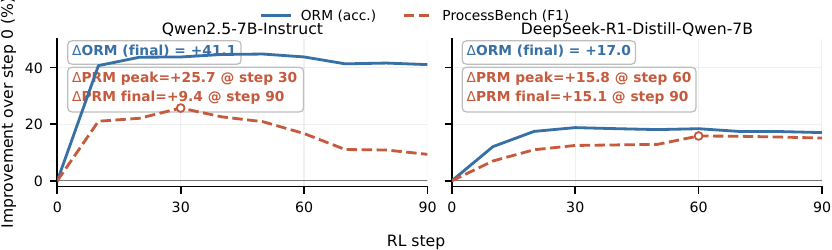}
    \caption{\textbf{Model priors affect outcome-to-process transfer.}
    Qwen2.5-7B-Instruct improves ORM rapidly but shows degraded ProcessBench F1
    at later checkpoints, while the DeepSeek-R1-distilled model maintains coupled
    ORM and PRM improvements.}
    \label{fig:orm_prm_decoupling}
\end{figure}

\paragraph{Model priors affect the stability of outcome-to-process transfer.}
To probe a possible failure mode of \method, we compare Qwen2.5-7B-Instruct with
its DeepSeek-R1-distilled counterpart. The results on ProcessBench are shown in Figure~\ref{fig:orm_prm_decoupling}. The distilled model shows
coupled improvement: both ORM accuracy and ProcessBench F1 increase and then
stabilize. In contrast, Qwen2.5-7B-Instruct improves ORM rapidly, but its
ProcessBench F1 peaks early and then deteriorates.
This suggests that \method\ relies on the model's initial prior. A weaker model may fit the outcome objective through a shortcut: it learns to predict whether the solution is ultimately wrong, while
its step-level judgments remain incomplete or weakly grounded.
The DeepSeek-distilled model, which has a stronger step-by-step
verification prior, instead converts outcome rewards into more stable process-level gains.
These findings suggest applying \method{} to backbones with strong CoT priors. 
Our observations also suggest a process-label-free early-stopping heuristic: stop training when ORM accuracy plateaus and response length shows a decline. This heuristic effectively help mitigate degradation in our experiments.

\begin{wraptable}[11]{r}{0.44\linewidth}
    \vspace{-8mm}
    \centering
    \small
    \setlength{\tabcolsep}{5pt}
    \renewcommand{\arraystretch}{1.08}
    \caption{\textbf{Prompting ablation on Qwen3-4B-Instruct-2507.}
    Values are measured at the end of the first epoch.}
    \label{tab:thought_vs_direct}
    \begin{tabular}{lcc}
        \toprule
        \textbf{Metric} & \textbf{CoT} & \textbf{Label-only} \\
        \midrule
        Base F1 & 63.8 & 47.5 \\
        PRM gain & 13.9 & 0.7 \\
        ORM gain & 17.6 & -5.6 \\
        Entropy & 0.31 & 0.002 \\
        Avg. length & 1828.7 & 47.9 \\
        Max KL loss & $1{\times}10^{-3}$ & $1.5{\times}10^{-5}$ \\
        \bottomrule
    \end{tabular}
    \vspace{-4mm}
\end{wraptable}

\paragraph{Thought is essential for PRM learning.}
To isolate the role of explicit reasoning, we replace CoT prompting with a
label-only prompt that directly outputs process and outcome labels. As shown in
Table~\ref{tab:thought_vs_direct}, the label-only variant has much lower base
ProcessBench F1 and gains little from outcome-supervised RL, while the CoT
variant achieves a large PRM improvement.
These results show that \method{} does not learn reliable process verification
from labels alone; the gains are mediated by reasoning traces that RL can shape.
The label-only variant's near-zero entropy, shorter outputs, and smaller KL loss
further suggest that it receives little effective learning signal.
\section{Conclusion}

We presented \method{}, an outcome-only RL framework that induces process
supervision through the chain-of-thought of a generative reward model. With only
3.2K outcome-labeled examples, \method{}-Qwen3-4B-Thinking-2507 achieves
\textbf{85.2 F1} on ProcessBench and improves AgentProcessBench FirstErrAcc
across all tested backbones. Our analyses show that CoT and the backbone's CoT
prior are key carriers of the induced signal, while weak outcome--process
coupling and outcome shortcuts limit the transfer. Future work may combine
\method{} with light Monte Carlo priors or CoT distillation.

\subsection*{AI use statement}
We used LLMs for language polishing, but not for generating research ideas or analyzing experimental results. 

\subsection*{Ethics statement}
We use open-weight policy models and publicly available datasets, and evaluate on public benchmarks. We do not foresee additional ethical concerns.

\subsection*{Reproducibility statement}
We detail the backbone models, datasets, and hyperparameters in Sections~\ref{sec:exp_setup} and Appendix~\ref{app:imp_details} to facilitate reproducibility.

\bibliographystyle{iclr2027_conference}
\bibliography{iclr2027_conference}

\appendix
\section{Limitations}\label{app:limit}

Our study has three main limitations. First, we focus on pure-text reasoning and
agent trajectories, and do not evaluate multimodal settings where process
verification must be grounded in images, videos, or other non-textual
observations. Second, \method{} relies on sufficient outcome--process coupling:
the induced process signal is weaker when final task success is only indirectly
related to local step correctness. Third, the
method depends on the backbone's ability to express verification behavior through
its generated thoughts. For weaker backbones or label-only prompting,
outcome-supervised RL may improve outcome prediction without reliably yielding
faithful step-level judgments.

\section{Broader Impacts}\label{boarder:impacts}

\method{} may reduce the cost of building process reward models by inducing
step-level verification from trajectory-level outcome labels, rather than
requiring human step annotations or extensive Monte-Carlo estimation. This can
benefit reasoning evaluation, educational feedback, and agent debugging, where
localizing errors is often more informative than judging only final outcomes.
However, stronger reward models may also increase the capability of systems
optimized with them, including potentially harmful or unsafe agents. Moreover,
because the process labels produced by \method{} are induced indirectly from
outcome rewards, they may be biased, miscalibrated, or unreliable in domains
where final success is weakly tied to local step correctness. These labels should
therefore not be treated as ground truth in high-stakes settings.
Responsible use of \method{} requires domain-specific validation, human
oversight, and separate evaluation of both outcome-level and step-level behavior.
Future work should study robustness, calibration, and misuse risks in broader
agentic and multimodal applications.

\section{GRPO Objective}\label{app:grpo}

For completeness, we provide the full GRPO objective used in our RL training.
For each input $(q,\mathbf{x})$, we sample a group of $G$ rollouts
$\{\mathbf{o}^{(1)},\dots,\mathbf{o}^{(G)}\}$ from the old policy
$\pi_{\theta_{\mathrm{old}}}(\cdot \mid q,\mathbf{x})$. Each rollout receives a
binary reward $r^{(i)}$ according to Eq.~\ref{eq:outcome_reward}. We then compute a
group-relative advantage as
\begin{equation}
    \hat{A}^{(i)} = r^{(i)} - \frac{1}{G}\sum_{j=1}^{G} r^{(j)} .
\end{equation}
Following recent practice~\citep{liu2025deepseek, liuunderstanding}, we use a
non-normalized variant of GRPO, i.e., we do not divide the advantage by the group
standard deviation.

The model is updated by maximizing the clipped surrogate objective:

\begingroup
\small
\begin{equation}
\begin{aligned}
\mathcal{J}_{\mathrm{GRPO}}(\theta)
=&\;
\mathbb{E}_{(q,\mathbf{x}) \sim \mathcal{D},\,
\{\mathbf{o}^{(i)}\}_{i=1}^{G} \sim
\pi_{\theta_{\mathrm{old}}}(\cdot \mid q,\mathbf{x})}
\Bigg[
\frac{1}{G}
\sum_{i=1}^{G}
\frac{1}{|\mathbf{o}^{(i)}|}
\sum_{t=1}^{|\mathbf{o}^{(i)}|}
\min \Bigg(
\rho_{t}^{(i)}(\theta)\hat{A}^{(i)},
\\
&
\mathrm{clip}\!\left(
\rho_{t}^{(i)}(\theta),
1-\epsilon,
1+\epsilon
\right)\hat{A}^{(i)}
\Bigg)
\Bigg],
\end{aligned}
\label{eq:grpo_objective}
\end{equation}
\endgroup
where

\begin{equation}
    \rho_{t}^{(i)}(\theta)
    =
    \frac{
    \pi_{\theta}(o_t^{(i)} \mid q,\mathbf{x},\mathbf{o}_{<t}^{(i)})
    }{
    \pi_{\theta_{\mathrm{old}}}(o_t^{(i)} \mid q,\mathbf{x},\mathbf{o}_{<t}^{(i)})
    },
\end{equation}
$|\mathbf{o}^{(i)}|$ is the length of the $i$-th rollout, and $\epsilon$ is the
clipping threshold.

\section{Derivations for Thinking-Induced Process Supervision}
\label{app:thinking_process_supervision}

This appendix provides proofs of Theorems~\ref{thm:outcome}
and~\ref{thm:process}.

\paragraph{Proof of Theorem~\ref{thm:outcome}.}
By the definition of conditional mutual information,
\begin{equation}
    I(Z;D \mid R)
    =
    H(D \mid R) - H(D \mid R,Z).
\end{equation}
It therefore suffices to upper-bound $H(D \mid R,Z)$.
Since $\hat D = g_\theta(R,Z)$ is a deterministic function of $(R,Z)$,
the pair $(R,\hat D)$ is also a deterministic function of $(R,Z)$.
Therefore,
\begin{equation}
    H(D \mid R,Z) \le H(D \mid R,\hat D).
\end{equation}
Moreover, conditioning cannot increase entropy, so
\begin{equation}
    H(D \mid R,\hat D) \le H(D \mid \hat D).
\end{equation}
Now $D$ and $\hat D$ are binary random variables. Applying Fano's
inequality with error probability
$P_e = \mathbb{P}[\hat D \neq D]$ gives
\begin{equation}
    H(D \mid \hat D)
    \le H_b(P_e) + P_e \log(|\{0,1\}|-1)
    = H_b(P_e).
\end{equation}
Combining the inequalities above,
\begin{equation}
    H(D \mid R,Z)
    \le H(D \mid R,\hat D)
    \le H(D \mid \hat D)
    \le H_b(P_e),
\end{equation}
and substituting back yields
\begin{equation}
    I(Z;D \mid R)
    =
    H(D \mid R) - H(D \mid R,Z)
    \ge
    H(D \mid R) - H_b(P_e). \qedhere
\end{equation}

\paragraph{Proof of Theorem~\ref{thm:process}.}
Apply the chain rule for conditional mutual information to
$I(Z;\mathbf Y,D \mid R)$ in two different orders. First,
\begin{equation}
    I(Z;\mathbf Y,D \mid R)
    =
    I(Z;\mathbf Y \mid R) + I(Z;D \mid \mathbf Y,R).
\end{equation}
Second,
\begin{equation}
    I(Z;\mathbf Y,D \mid R)
    =
    I(Z;D \mid R) + I(Z;\mathbf Y \mid D,R).
\end{equation}
Equating the right-hand sides and rearranging gives
\begin{equation}
    I(Z;\mathbf Y \mid R)
    =
    I(Z;D \mid R)
    + I(Z;\mathbf Y \mid D,R)
    - I(Z;D \mid \mathbf Y,R).
\end{equation}
Since mutual information is nonnegative,
\begin{equation}
    I(Z;\mathbf Y \mid R)
    \ge
    I(Z;D \mid R) - I(Z;D \mid \mathbf Y,R).
\end{equation}
Finally,
\begin{equation}
    I(Z;D \mid \mathbf Y,R)
    =
    H(D \mid \mathbf Y,R) - H(D \mid \mathbf Y,R,Z)
    \le
    H(D \mid \mathbf Y,R),
\end{equation}
because entropy is nonnegative. Substituting yields
\begin{equation}
    I(Z;\mathbf Y \mid R)
    \ge
    I(Z;D \mid R) - H(D \mid \mathbf Y,R). \qedhere
\end{equation}



\section{Implementation Details.}\label{app:imp_details}
For the math domain, we sample 3,200 trajectories from the published SCAN-Pro\footnote{\url{https://huggingface.co/datasets/dyyyyyyyy/SCAN-Pro}} dataset, which is constructed from the MATH training set. For the agent domain, we construct an RL training corpus of 2,905 trajectories from HotpotQA~\cite{yang2018hotpotqa}, GAIA~\cite{mialon2023gaia}, $\tau^2$-Bench~\cite{barres2025tau2}, and BFCL~\cite{patil2025the}.
To mitigate test-set leakage, we decontaminate the training corpus against AgentProcessBench by excluding trajectories with task-prompt 5-gram Jaccard similarity above 0.8.
All experiments are conducted on 8 NVIDIA A800 GPUs with 80GB memory each.
We conduct RL training with the \texttt{verl} framework. We set the maximum rollout response length to 16K tokens, set the KL penalty coefficient to 0, train for 1 epoch with AdamW, and use batch size of 32 and a learning rate of $1\times10^{-6}$.

\section{Prompt Design}
\label{app:prompt_design}

In this section, we present the used prompt in \method{}.
\begin{tcolorbox}[
    breakable,
    title={Prompt for Mathematical Reasoning},
    fonttitle=\bfseries,
    fontupper=\small,
    colback=white,
    colframe=black!60,
    colbacktitle=black!6,
    coltitle=black,
    boxrule=0.5pt,
    arc=1mm,
    left=8pt,
    right=8pt,
    top=6pt,
    bottom=6pt
]
The following is a math problem and a solution split into paragraphs and indexed from 0.

\medskip
\noindent\textbf{[Math Problem]}\\
\texttt{PROBLEM}

\medskip
\noindent\textbf{[Solution]}\\
\texttt{TAGGED\_RESPONSE}

\medskip
Your task is to judge the solution paragraph by paragraph.

First, think through the solution carefully and identify where the reasoning first becomes invalid.

Then output a JSON object with:
\begin{itemize}[leftmargin=1.5em, topsep=3pt, itemsep=3pt, parsep=0pt]
    \item \texttt{step\_labels}: a mapping from paragraph index to 1 or 0.
    \begin{itemize}[leftmargin=1.5em, topsep=2pt, itemsep=2pt, parsep=0pt]
        \item 1 means the paragraph is correct and remains logically valid given the previous paragraphs.
        \item 0 means the paragraph is incorrect, unsupported, or depends on an earlier incorrect paragraph.
    \end{itemize}
    \item \texttt{final\_label}: 1 if the final answer of the whole solution is correct, otherwise 0.
\end{itemize}

\textbf{Rules:}
\begin{itemize}[leftmargin=1.5em, topsep=3pt, itemsep=2pt, parsep=0pt]
    \item You must reason before the final JSON.
    \item \texttt{step\_labels} must contain every paragraph index exactly once.
    \item Use only 1 or 0 for every label.
    \item If an earlier paragraph is wrong and a later paragraph relies on it, that later paragraph should also be 0.
    \item The final output must end with a \verb|```json| code block containing only the final JSON object.
\end{itemize}

\textbf{Illustrative output format:}
\begin{verbatim}
```json
{
  "step_labels": {"0": 1, "1": 1, "2": 0},
  "final_label": 0
}
```
\end{verbatim}
\end{tcolorbox}

\section{Human Evaluation of Process-Level Feedback}
\label{app:human_evaluation}
To assess the quality of the process-level labels generated by TIPS, we sampled 60 trajectories from ProcessBench, with 15 trajectories from each subset. Two PhD-level annotators evaluated the feedback generated by TIPS-Qwen3-4B-Instruct-2507. The evaluation examined whether the model correctly localized the first error and whether its accompanying rationale accurately explained the underlying error.
The model correctly localized the first error in 47/60 cases (78.3\%). Among these correctly localized cases, the rationale accurately explained the underlying error in 38/47 cases (80.9\%). Thus, 38/60 cases (63.3\%) exhibited both correct error localization and a human-validated explanation. These results provide additional evidence that \method{} can generate process-level feedback that both identifies and explains reasoning errors.

\section{Does \method{} Simply Improve Problem-Solving Ability?}
\label{app:solver_verifier}

To investigate whether the improvements from \method{} can be explained by stronger mathematical problem solving, we conduct two complementary analyses. First, we evaluate how \method{} training affects the direct problem-solving performance of its backbones. Second, we compare \method{} with a solver-format RL baseline and additional models with higher solver benchmark scores. We evaluate direct problem solving using mean@4 accuracy on AIME 2026 and HMMT February 2026, and process verification using F1 on ProcessBench.

\begin{table}[t]
    \centering
    \small
    \setlength{\tabcolsep}{4pt}
    \renewcommand{\arraystretch}{1.08}
    \caption{\textbf{Direct problem solving and process verification.} AIME 2026 and HMMT February 2026 scores are mean@4 accuracy (\%). ProcessBench scores are evaluated in F1. The solver RLVR baseline is initialized from Qwen3-4B-Instruct-2507 and trained on DAPO-Math-17K. }
    \label{tab:solver_verifier_comparison}
    \begin{tabular}{@{}lccc@{}}
        \toprule
        \textbf{Model} & \textbf{AIME 2026} & \textbf{HMMT Feb.\ 2026} & \textbf{ProcessBench} \\
        \midrule
        Qwen3-4B-Instruct-2507 & 53.3 & \textbf{36.4} & 63.8 \\
        \quad + Solver RLVR & \underline{55.8} & \underline{33.3} & 62.8 \\
        \quad + TIPS & 53.3 & 31.8 & \textbf{77.7} \\
        GooseReason-4B-Instruct & \textbf{62.5} & \underline{33.3} & \underline{72.7} \\
        \midrule
        Qwen3-4B-Thinking-2507 & 80.0 & 53.0 & 60.3 \\
        \quad + TIPS & \underline{80.8} & \underline{54.6} & \textbf{85.2} \\
        Qwen3-30B-A3B-Thinking-2507 & \textbf{86.7} & \textbf{74.2} & \underline{80.3} \\
        \bottomrule
    \end{tabular}
\end{table}

\paragraph{Direct problem-solving performance.}
As shown in Table~\ref{tab:solver_verifier_comparison}, \method{} improves ProcessBench F1 by 24.9 points for Qwen3-4B-Thinking-2507 and 13.9 points for Qwen3-4B-Instruct-2507. These improvements are accompanied by much smaller changes in direct problem solving: the thinking variant improves modestly on both solver benchmarks, whereas the Instruct variant remains unchanged on AIME 2026 and declines on HMMT February 2026. Thus, \textbf{the substantial gains in process verification are not accompanied by comparable improvements on the two solver benchmarks}.

\paragraph{Comparison with solver and verification baselines.}
We train a same-backbone control using standard solver-format reinforcement learning with verifiable rewards (RLVR) on DAPO-Math-17K, with questions as inputs and final-answer correctness as the reward. This baseline scores higher than TIPS-Qwen3-4B-Instruct-2507 on both AIME 2026 and HMMT February 2026, but trails it by 14.9 F1 points on ProcessBench. Therefore, the solver-format RLVR baseline does not reproduce the process-verification gains achieved by \method{}, despite its higher direct problem-solving scores.

The additional comparisons show a similar pattern. GooseReason-4B-Instruct outperforms TIPS-Qwen3-4B-Instruct-2507 on both solver benchmarks but trails it by 5.0 F1 points on ProcessBench. Likewise, Qwen3-30B-A3B-Thinking-2507 achieves higher solver scores than TIPS-Qwen3-4B-Thinking-2507 while scoring 4.9 F1 points lower on ProcessBench. These comparisons demonstrate that, among the evaluated models, \textbf{higher solver benchmark scores do not necessarily translate into better step-error localization}. Together, these findings support the interpretation that \method{} strengthens process verification beyond the improvements observed in direct problem solving.

\end{document}